\documentclass{article}
\usepackage{ijcai24}

\usepackage{times}
\usepackage{soul}
\usepackage{url}
\usepackage[hidelinks]{hyperref}
\usepackage[utf8]{inputenc}
\usepackage[small]{caption}
\usepackage{graphicx}
\usepackage{amsmath}
\usepackage{amsthm}
\usepackage{booktabs}
\usepackage{algorithm}
\usepackage{algorithmic}
\usepackage[switch]{lineno}

\usepackage{xcolor}
\usepackage{amssymb}
\usepackage{soul}
\usepackage{adjustbox}

\usepackage{caption}
\usepackage{subcaption}

\newcommand{\norm}[2]{\left \lVert #1 \right \rVert_{#2}}

\title{LLM-assisted Concept Discovery: Automatically Identifying and Explaining Neuron Functions}

\author{
Xuan-Nhat Hoang$^1$
\and
Minh Vu$^1$
\and
My T. Thai$^1$
\affiliations
$^1$University of Florida
\emails
\{hoangx, minhvu\}@ufl.edu,
mythai@cise.ufl.edu
}

\begin{document}

\maketitle

\begin{abstract}
    Providing textual concept-based explanations for neurons in deep neural networks (DNNs) is of importance in understanding how a DNN model works. Prior works have associated concepts with neurons based on examples of concepts or a pre-defined set of concepts, thus limiting possible explanations to what the user expects, especially in discovering new concepts. Furthermore, defining the set of concepts requires manual work from the user, either by directly specifying them or collecting examples. To overcome these, we propose to leverage multimodal large language models for automatic and open-ended concept discovery. We show that, without a restricted set of pre-defined concepts, our method gives rise to novel interpretable concepts that are more faithful to the model's behavior. To quantify this, we validate each concept by generating examples and counterexamples and evaluating the neuron's response on this new set of images. Collectively, our method can discover concepts and simultaneously validate them, providing a credible automated tool to explain deep neural networks. 

\end{abstract}

\section{Introduction}
\label{sec:introduction}

As large deep neural networks become more prevalent in everyday life, so does the need to understand their decision-making process to ensure their trustworthiness. Careful analysis of deep neural networks (DNNs) by looking at individual neurons and their combinations, has yielded valuable insights on how DNNs function \cite{cammarata_thread_2020,goh_multimodal_2021,bricken_towards_2023}. For example, in the GPT-2 model, a specific set of neurons has been discovered to perform the natural language task of indirect object identification \cite{wang_interpretability_2022}. In the vision domain, some neurons have been found to detect curves \cite{cammarata_curve_2020}. However, these analyses require extensive human labor in examining the neurons and coming up with possible explanations.

Concept-based explanations seek to characterize the model's global behavior via concepts, which are related to how human reasons \cite{yeh_completeness-aware_2020}. These concepts serve to provide an intuition on the common characteristics of the inputs that the network recognizes. They can be more intuitive and require less time to inspect than individual inputs.

In providing concept-based explanations to neurons in DNNs, we seek to characterize each neuron by a textual concept that represents a common trait of the inputs activating this neuron. Moreover, we aim to do this automatically, reducing the need for human interference in defining the concepts and validating the generated output. Recently, Large Language Models (LLMs) have been leveraged to automatically generate explanations for other language models\cite{bills_language_2023,singh_explaining_2023}. LLMs are further utilized to score the resulting explanations, creating a complete framework.
% \st{In the language domain, leveraging the advances of powerful large language models (LLMs), emerging methods \cite{gpt3explaingpt2, Microsoft saas} have used them to automatically generate explanations for other models. The same LLM can also be used to score the resulting explanations, creating a complete framework.} 
It is natural to pose the question of whether a textual concept-based explanation method with the same degree of automation can be realized in the vision domain. In fact, this work demonstrates that, by utilizing LLMs, we can overcome the limitations of existing works \cite{hernandez_natural_2022,kalibhat_identifying_2023} in terms of the explainable concepts' space and the explaining language. Those limitations come from the fact that the existing explaining algorithms strongly depend on human annotations or curation of a large image-text dataset.
% \st{Existing works~\cite{hernandez_natural_2022,kalibhat_identifying_2023} require definitions of concepts at the training/pre-processing time, either via human annotation or curation of a large image-text dataset. In turn, this makes their outputs dependent on the training data, both in terms of the space of explainable concepts and the language of the output.}

% Since LLMs are trained for language understanding, they have the added benefit of being able to present the explanations in human-understandable language.

% Scaling up such analysis can shed light on the decision-making process of neural networks, leading to various application such as auditing or intervention \cite{gan_dissect,rome}. \mt{Should say how important the concept explanation is}

% To assist with the formalization of such hypotheses, automated interpretability methods can be utilized. Many existing explanation methods yield outputs in the form of a heat map \cite{grad-cam}, exemplifying images \cite{feature_viz}, or prototypical vectors \cite{tcav, ace}. These methods are useful, however, it is challenging to automatically interpret their outputs.  \mt{These are not concept explanation so they're not related. save space. Not even need to discuss them. Only discuss related concept-explanation on neuron's work}

In particular, we propose a method to automatically discover concepts for neural representations. Our method probes the neuron with an image-only dataset, identifies an interpretable concept among the highly activating images, and explains it using a Multimodal Large Language Model (MLLM). Our method is training-free, does not require examples of a concept, and can recognize concepts in an open-ended manner. Beyond concept discovery, we propose a procedure to evaluate a proposed concept independently of the probing dataset. Our score compares the activation of a neuron on examples possessing the concept and non-examples containing the concept's co-hyponyms. This serves to quantify the ability of the neuron in telling the concept from other semantically similar concepts. Intuitively, given a concept produced by our concept discovery method, these co-hyponyms represent the concepts one would test the neuron on to ensure the specificity of the proposed concept, therefore, our method helps with automating this process. To verify the effectiveness of our method, we apply our approach in examining various pre-trained models and demonstrate how improved explanations can help with scaling up the analysis of neural networks.

% We adopt a multimodal large language models (MLLM) to automatically recognize open-ended visual concepts, without the need for labeled data. The ability of MLLMs to answer visual questions is directly used to extract the concepts in textual format.

The rest of this paper is organized as follows. Section~\ref{sec:related_work} discusses the related work and some preliminaries. Section~\ref{sec:concept_discovery} describes our concept discovery method. Section~\ref{sec:concept_validation} shows how we compute our validation score. We conduct experiments to show the effectiveness of our method in Section~\ref{sec:experiments}, and finally Section~\ref{sec:conclusion} presents our conclusion.

\section{Preliminary and Related Work}
\label{sec:related_work}

\paragraph{Visual neuron-concept association }

Network Dissection \cite{bau_network_2017} proposes to label neurons with textual concepts by computing the similarity between the output feature map and the output of a segmentation model. Since then, many methods \cite{mu_compositional_2020,oikarinen_clip-dissect_2023,bykov_labeling_2023} have been developed to associate concepts with neurons using different criteria. Since they only focus on matching the concepts, they usually require a pre-defined set of concepts as input or default to simple alternatives (e.g., 20,000 most common English words \cite{oikarinen_clip-dissect_2023}). As such, these methods rely on the user anticipating the possible concepts, or their vocabulary will be severely limited. Because the association methods only work when comparing different concepts, in our work, we propose an example-based causality criterion to evaluate a single concept that can also be used for neuron-concept association.

\paragraph{Automatic vision concept discovery } Unlike concept association, where concepts are provided, concept discovery needs need to come up with a concept from the vocabulary. FALCON \cite{kalibhat_identifying_2023}  addresses the challenge of a limited concept set by extracting terms from a large image-text dataset like LAION-400M \cite{schuhmann_laion-400m_2021}.  While they benefit from the variety of concepts, they can only output words and phrases present in the dataset, greatly limiting the flexibility of their output. On the other hand, MILAN \cite{hernandez_natural_2022} is the first generative method that can generalize to different domains and tasks. However, the method requires a specific dataset of images with masks annotated concepts, posing challenges in obtaining such annotated data. Our method exploits the inherent capabilities of MLLMs and uses them for \textit{concept discovery}, eliminating the need for annotated data. As MLLMs improve across domains, our method will directly inherit those improvements, while previous works will require additional collection of data.

% On the other hand, MILAN, proposed by Hernandez et al. (2022), is the first to present a generative model capable of generalizing to various domains and tasks. However, MILAN requires masks annotated with concepts for training, posing challenges in obtaining such annotated data.

% Several techniques for automatic concept discovery at the \textit{model level} (not neuron-level) have been developed, such as ACE, ConceptShap, Feature Viz, Hernandez et al. (2022), and Kalibhat et al. (2023) \cite{ace, conceptshap, feature_viz, hernandez2022natural, kalibhat_identifying_2023}. However, the results of these methods are images or vectors, necessitating manual effort for further interpretation. In this work, our focus is to return natural language explanations \cite{hernandez_natural_2022, kalibhat_identifying_2023}.

% 

% Our method exploits the inherent capabilities of MLLMs, utilizing them for concept discovery without the need for annotated data. As MLLMs progress across domains, our approach directly inherits those improvements, eliminating the necessity for additional data collection compared to previous methods.

\paragraph{Multimodal large language models} MLLMs are similar to LLMs in the sense that they both accept and yield text, however, MLLMs can accept an image as additional context. MLLMs can connect information present in images and text, allowing it to answer questions such as ``What does this image describe?" 

Available MLLMs generally belong to one of two categories: open-source MLLMs and proprietary MLLMs. While open-source MLLMs allow the fine-tuning of MLLMs for a specific task, proprietary MLLMs are often much larger, more capable, and easier to integrate, only requiring an external API call. We elect to use a proprietary MLLM to ensure the highest quality for the task of concept discovery. We compare with open-source MLLMs in ablation tests in the supplementary material.

% \paragraph{Natural language explanations of visual features } 
% Natural language enables the conceptualization of more complex ideas, such as logical forms \cite{CompExp}, that are otherwise hard to express with images alone. Previous natural-language based methods \cite{netdissect, clip-dissect, invert} are usually limited to a set of pre-defined concepts, limiting their adaptability to new settings. 

\section{LLM-assisted Concept Discovery} 
\label{sec:concept_discovery}

\begin{figure*}
    \centering
    \includegraphics[width=0.9\textwidth]{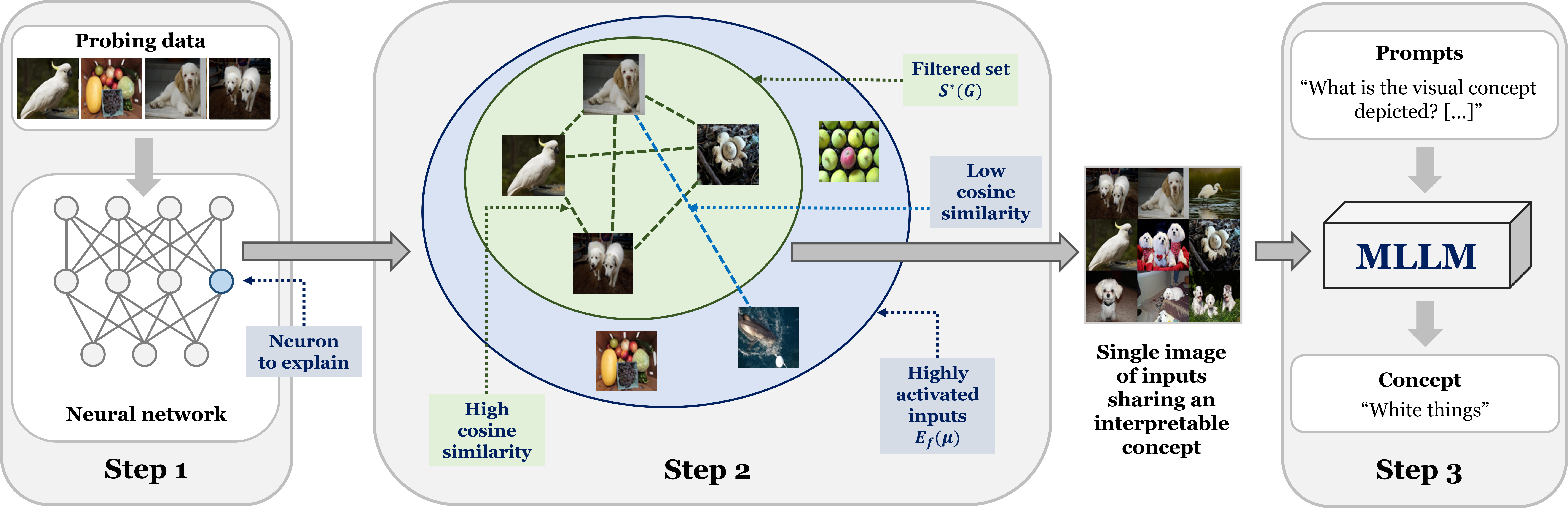}
    \caption{Overview of our LLM-assisted concept discovery algorithm. Given a neuron of interest, we generate an exemplar representation from the probing dataset. Next, we find an interpretable concept by looking for a group of images with high average CLIP similarity. Finally, we prompt the MLLM with the combined image for the common concept that the neuron activates on.}
    \label{fig:concept_discovery}
\end{figure*}

In this section, we first describe our Concept Discovery problem in Section ~\ref{sec:sub:formulation}. Since it is infeasible to directly solve that problem, we introduce the LLM-assisted Concept Discovery algorithm that, given a neuron, automatically generates a concept $c$ that the neuron captures. The algorithm consists of 3 main steps, which are illustrated in Fig.~\ref{fig:concept_discovery}. Sect.~\ref{sec:sub:exemplar} describes the first step selecting a set of inputs that highly activate the neuron, called the \textit{exemplar representations} $E_f(\mu)$. To ensure the algorithm returns a single interpretable concept, the second step (Sect.~\ref{sec:sub:finding}) eliminates inputs with low cosine similarity from $E_f(\mu)$ and returns a more filtered set. Given that, the final step of the algorithm utilizes a MLLM to generate a concept explaining the examined neuron (Sect.~\ref{sec:sub:mllm_propose}).

% Our algorithm takes in a probing dataset and a neuron to be explained and generates a concept explanation. Figure~\ref{fig:concept_discovery} illustrates the main steps of our algorithm. Then, Sections~\ref{sec:sub:exemplar}, \ref{sec:sub:finding} and \ref{sec:sub:mllm_propose} discuss in detail each main step of our solution.

\subsection{Problem formulation} \label{sec:sub:formulation} 
We consider a  neural network $\Phi:\mathcal{X}\rightarrow \mathcal{Y}$, where $\mathcal{X}$ and $\mathcal{Y}$ denote the input and output domains, respectively. Given an arbitrary neuron in $\Phi$, we refer to it by its activation function $f: \mathcal{X} \rightarrow \mathbb{R}$. Given a dataset $D\subset \mathcal{X}^{*}$ and a concept $c$, we define $D_c$ and $D_{\Bar{c}}$ the set of $D$'s elements that contains and does not contain $c$, respectively:
\begin{align*}
    D_{c} := \{x\in D | \mathcal{O}_c(x) = 1 \} \\
    D_{\Bar{c}} := \{x\in D | \mathcal{O}_c(x) = 0 \}
\end{align*}
where $\mathcal{O}_c(x)$ is an oracle determining whether or not the concept $c$ occurs in $x$, i.e., $\mathcal{O}_c(x) = 1$ %if and only if
iff $c$ appears in $x$.

Given a neuron or, equivalently, its activation $f$, the goal of this work is to find a descriptive concept $c$ that the neuron captures: we say a neuron $f$ captures a concept $c$ if $f$ can differentiate the elements of $D_{c}$ from the elements of $D_{\Bar{c}}$. More formally, we aim to generate the concept that maximizes:
\begin{align}
    c^* = \operatorname*{arg\,max}_{c} \textup{Pr} [f(X_1) > f(X_2) | X_1 \in D_c, X_2 \in D_{\Bar{c}}] \label{eq:obj}
\end{align}
In other words, we aim to find the concept $c$ such that the activation of a random input $X_1$ sampled from $D_c$ has a high probability of being greater than that of an $X_2$ sampled from $D_{\Bar{c}}$. The intuition for this objective is, for a solution $c$ of (\ref{eq:obj}), we can expect that $f(x)$ for an $x \in D_c$ is generally greater than $f(x), x\in D_{\Bar{c}}$. It implies that we can directly use $f(x)$ to differentiate $D_c$ from $D_{\Bar{c}}$, i.e., $f(x)$ indeed capture the concept $c$. 

There are several challenges to directly solve (\ref{eq:obj}). First, there is no trivial way to search, iterate, and/or optimize over the solution's space, i.e., the space of concepts, especially when it might not even be a metric space. Secondly, even if we can construct a metric space on the set of concepts, computing the actual value of (\ref{eq:obj}) is not trivial. The reasons are the absence of the oracle $\mathcal{O}$ in practice and complex behavior $f$, which prevent efficient computations of the probability.

% Instead of directly solving (\ref{eq:obj}), we introduce the LLM-assisted Concept Discovery algorithm that, given a neuron $f$, automatically generates a concept $c$ that the neuron $f$ captures. The algorithm consists of 3 main steps (Fig.~\ref{fig:concept_discovery}). The first selects a set of inputs that highly activate the neuron $f$, called the \textit{exemplar representations} $E_f(\mu)$. To ensure the algorithm returns a single interpretable concept, the second step eliminates inputs with low cosine similarity from $E_f(\mu)$. Given that set, the final step of the algorithm utilizes a MLLM to generate a concept explaining the neuron $f$.

% \mv{include pseudo-code, high-level description, and figure (if needed) to illustrate the algorithm.}

% \mt{Yes, refer to fig 1. to give an abstract level discussion of the overall solution}

\subsection{Exemplar representation}
\label{sec:sub:exemplar}

The first step of the algorithm is to generate a set of \textit{exemplar representations}~\cite{hernandez_natural_2022}, which is given as:
\begin{align}
    E_f(\mu) := \{x\in X | f(x) \geq \mu\}
\end{align}
where $\mu$ is a threshold parameter. Intuitively, $E_f(\mu)$ is the set of inputs that highly activate the neuron's activation, i.e., the inputs whose activation at the neuron $f$ is not lower than $\mu$.

Although the exemplar representations $E_f(\mu)$ may not fully characterize the neuron, previous studies have demonstrated that meaningful information can still be extracted from this set alone. For example, \cite{bau_understanding_2020,bau_gan_2019} utilize it to identify the neurons that trigger certain class predictions or generate specific objects. \cite{andreas_learning_2018}~employ these sets to forecast output sequence for newly observed inputs. Additionally, ~\cite{olah_building_2018} and~\cite{mu_compositional_2020} use them to pinpoint adversarial vulnerabilities. Following previous research, the first step of our proposed concept discovery algorithm also represents the examined neuron as the set of exemplar representations. In particular, we use a single dataset $D_{probe}$ consisting only of images. For all neurons we wish to explain in a model, we record the neuron activations on this dataset and extract the exemplar representation from the top activating images of this dataset. In the next sections, we explain how we utilize a MLLM to extract the desired concept from the exemplar representation.

\subsection{Filtering the exemplar representation for interpretable concept identification}
\label{sec:sub:finding}

Given an exemplar representation, we wish to extract a concept that is common to this set of images. However, it is natural to pose the question whether such a concept always exists. Neurons are known to be \textit{polysemantic}, whereby a single neuron can respond to multiple unrelated concepts~\cite{omahony_disentangling_2023,olah_zoom_2020}. This polysemantic character of a neuron is expected to appear and influence the exemplar representations, making it challenging to identify a single concept explaining the neuron. Secondly, when the neuron exhibits polysemy, the MLLM might fail to generate a meaningful output, whose reason is discussed later in Section~\ref{sec:sub:mllm_propose}.

To address those issues, prior to extracting, the second step of our algorithm involves the selection of a subset of $E_f{(\mu)}$ that promotes the discovery of a single interpretable concept. Specifically, we adopt the observation of~\cite{kalibhat_identifying_2023} that members of $E_f{(\mu)}$ with high average cosine similarity are more likely to correspond to a single interpretable semantic concept. As such, we consider the problem of determining this subset as a problem of finding a subgraph with minimal diameter. Let $h$ denote the CLIP vision encoder. We construct a weighted complete graph $G = (V, E)$ where $V \equiv E_f(\mu)$ and a distance function $d: E \rightarrow \mathbb{R}$ such that $d(u, v) = \norm{h(u) - h(v)}{2}$. The $M$ images to be kept are those in the following solution:
\begin{align}
\label{eq:clip_sampling}
    S^*(G) = \operatorname*{arg\,min}_{S \subset V, |S|=M} \max\{d(u, v)|u,v \in  S\} 
\end{align} 
This problem is reducible to the clique decision problem, hence it is NP-complete. However, since $M$ and $|V|$ are small, the solution can be found in a short time with binary search and an efficient clique-finding routine. This subset is then used to prompt the MLLM, as further described in Section~\ref{sec:sub:mllm_propose}.

\subsection{MLLM as concept proposer}
\label{sec:sub:mllm_propose}

Directly extracting the concept $c$ from the filtered exemplar representation is still non-trivial due to the modality gap. This problem can be viewed as a summarization task, as we wish to concisely describe the common feature of a set of images with only a short phrase. However, this \textit{multi-image summarization} task is less explored compared to its natural language counterpart, the multi-document summarization, due to the scarcity of related datasets and models~\cite{trieu_multi-image_2020}.  MILAN \cite{hernandez_natural_2022} tackled the initial challenge by designing a specific dataset for neuron explanations and training the explanation model. However, adopting the method to another domain requires the collection of new data with specialized labels, which can be costly and/or even infeasible in many applications. In contrast, our work leverages the recent advancement of MLLM and utilizes it as a concept proposer.

We utilize an API-based MLLM (GPT-4V), as it is more capable than open-source alternatives. Note that, since we have no access to the intermediate representations of the inputs nor their gradients, the task needs to be fully specified by the inputs alone. To this end, we pose the problem as a Visual Question Answering (VQA) task where we give the MLLM images and ask for a common visual concept. The MLLM that we use can take multiple images, however, its ability to comprehend dozens of images is not well-studied. Furthermore, doing so would incur tremendous API costs, reducing the utility of our method. To make our method adaptable to the majority of MLLMs which can take only one image and to reduce API costs, our algorithm transforms $E_f(\mu)$ into a singular image while preserving the needed information. This is achieved by downsampling the images and arranging them in a grid-like pattern to form a single image, as shown in Figure \ref{fig:concept_discovery}. As shown by previous work~\cite{dunlap_describing_2023}, MLLMs are capable of understanding this kind of ``condensed" image. 

Despite the MLLM's great capabilities, we observed that it can produce unwanted outputs. Particularly, it can produce concepts that are too abstract or unhelpful for the task (for example, ``this image features a variety of objects in different settings)." Since we cannot ``train" the model, we try to guide the model towards desired outputs by putting additional information in the prompt. For LLMs, the common method is to perform In-context learning, where examples of inputs and answers are presented in the prompt so the LLM can ``learn" from them. In our case, doing so necessitates multiple image inputs, which would dramatically raise the API costs. We opt for a simpler alternative:  we collect unwanted outputs during initial trials and use them as examples of bad answers in the final prompt. This strategy reduces those unwanted concepts to some extent, while keeping the input length and costs reasonable.

% Another obstacle to our usage of MLLMs is the built-in safety mechanisms. While general-purpose MLLMs possess general knowledge that is needed for automatic novel concept discovery, they are also trained to refuse certain kinds of prompts \cite{openai_gpt-4_2023}. For the MLLM that we deploy in our method (GPT-4V), the prompt characteristics that cause the model to refuse the request have not been provided. However, during our trials, we found that the few cases which GPT-4V refused to answer are usually cases where the exemplar representation does not appear to contain a single identifiable common concept. Thus, we hypothesize that the model will refuse to answer when it is not confident of the answer, likely in an attempt to reduce misinformation. This motivates us to have an additional processing step before prompting the MLLM, which we discuss in the section below. 

Another challenge in utilizing MLLMs is the presence of built-in safety mechanisms. While general-purpose MLLMs possess the necessary general knowledge for automatic novel concept discovery, they are also designed to reject certain types of prompts \cite{openai_gpt-4_2023}. In the case of the MLLM deployed in our method (GPT-4V), the prompt characteristics leading to model refusal have not been explicitly provided. However, through our trials, we observed that GPT-4V tends to refuse instances in which the exemplar representation does not appear to contain a single identifiable common concept. Consequently, we hypothesize that the model rejects responses when it lacks confidence, possibly as a measure to mitigate misinformation. This motivates us to have an additional processing step before prompting the MLLM, which is discussed in the section below.

\section{Concept validation}
\label{sec:concept_validation}

\begin{figure}
    \centering
    \includegraphics[width=0.482\textwidth]{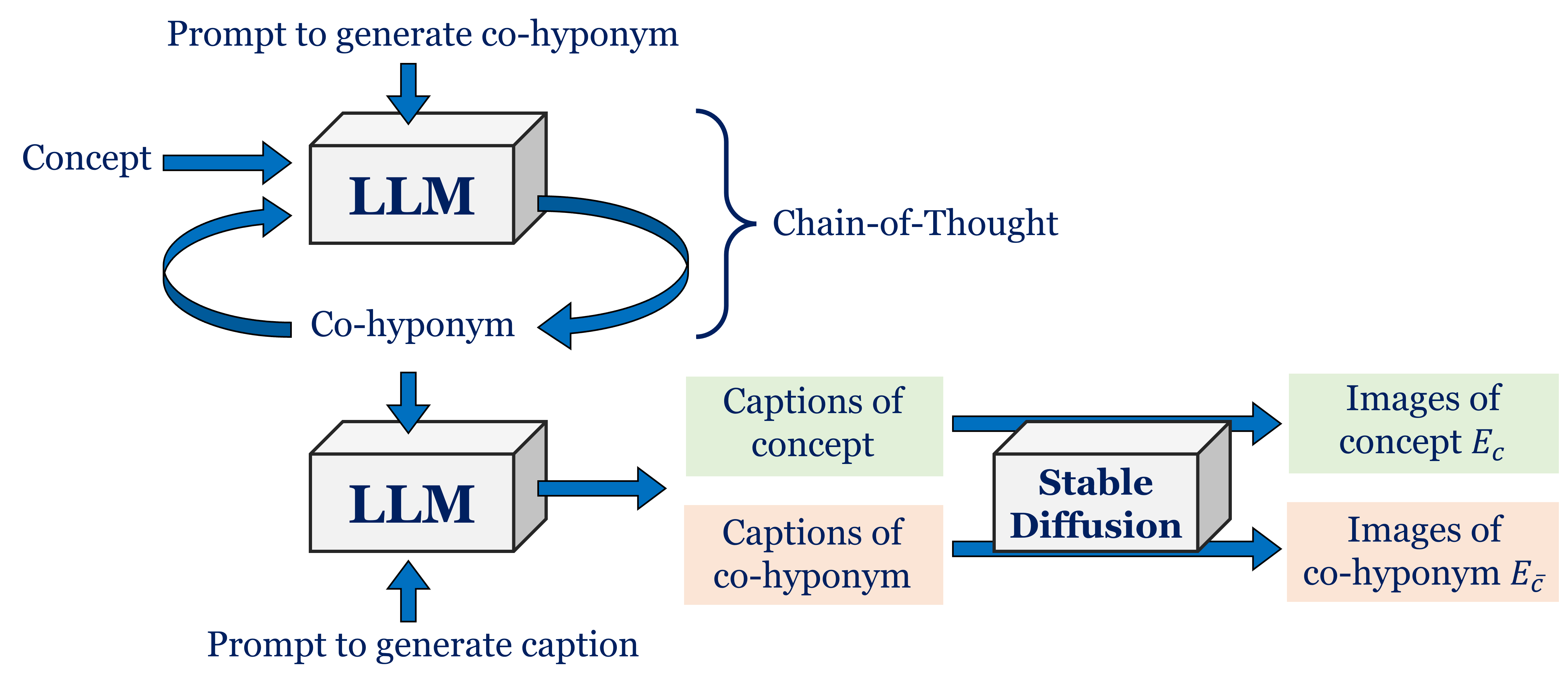}
    \caption{Overview of our concept evaluation. Given a concept, we use  Chain-of-Thought prompting on an LLM to generate its co-hyponyms. Then, we use another prompt to generate captions for the concept and the co-hyponyms. The last step utilizes a diffusion model to transform the caption into the sets of examples and non-examples $E_c$ and $E_{\bar{c}}$}
    \label{fig:validation}
\end{figure}

Beyond concept discovery, in this paper, we propose a score to validate the discovered concept. Our evaluation takes in a neuron and its proposed concept $c$, and returns a score which can be used to quantify whether the neuron's activation is consistently higher for inputs containing $c$. The overview of our evaluation is depicted in Fig.~\ref{fig:validation}.

The following sections are organized as follows. In Section~\ref{sec:sub:score_formulation}, we formulate our scoring objective and relate it to our previous goal of concept discovery in (\ref{eq:obj}). Section~\ref{sec:sub:sampling_hard} describes our proposed idea of finding harder non-examples to better evaluate the proposed concept. Section~\ref{sec:sub:finding_co_hyponyms} and Section~\ref{sec:sub:concept_to_image} discuss how our evaluation can be realized.

\subsection{Objective formulation}
\label{sec:sub:score_formulation}

Since the objective of concept discovery is (\ref{eq:obj}), it is intuitive to use the following score to evaluate our solution:
\begin{align}
\label{eq:score_validation}
    s(c) = \textup{Pr} [f(X_1) > f(X_2) | X_1 \in D_c, X_2 \in D_{\Bar{c}}]
\end{align}
As mentioned previously in Section~\ref{sec:sub:formulation}, it is non-trivial to compute this score due to the absence of the oracle $\mathcal{O}$. Previous approach~\cite{bykov_labeling_2023} overcomes this issue by relying on the availability of concept labels along with the dataset, which may not be the case in many scenarios. Even if we could approximate the oracle $\mathcal{O}$, finding a comprehensive dataset $D$ such that $D_c \neq \emptyset$ for all concept $c$ learned by a neural network is non-trivial. To address this issue, we propose a generative approach to obtain sets of examples and non-examples, $E_c$ and $E_{\bar{c}}$, as substitutes for $D_c$ and $D_{\bar{c}}$ in the evaluation (\ref{eq:score_validation}). Given $E_c$ and $E_{\bar{c}}$, we compute the following score:
\begin{align}
\label{eq:score_auc}
    s(c) = \frac{\sum_{x_1 \in E_c} \sum_{x_2 \in E_{\bar{c}}} 1[f(x_1) > f(x_2)]}{|E_c| \cdot |E_{\bar{c}}| }
\end{align}

The intuition is the same as discussed in Section~\ref{sec:sub:formulation}: when we say a neuron activates on concept $c$, we expect the value of $f(x), x \in E_c$ is higher than $f(x), x \in E_{\bar{c}}$. This score is closely related to the Area Under the Receiver Operating Characteristic curve (AUC), a metric used to evaluate binary classifiers, however, they are different due to our definition of $E_c$ and $E_{\bar{c}}$.

Our validation process consists of four steps. Given a concept $c$, the first step is to generate the set of co-hyponyms of $c$, whose definition we will give in the next sub-section. Next, we generate captions of images that contain $c$ and its co-hyponyms. Then, the captions are used to generate the sets of examples and non-examples $E_c$ and $E_{\bar{c}}$. Finally, those sets are used to compute the score in Equation (\ref{eq:score_auc}).

\subsubsection{Definitions}

Before we further describe how we generate $E_c$ and $E_{\bar{c}}$, we state some definitions that we will make use of extensively in the later sections.

\paragraph{Hypernym} A hypernym of a concept is another concept that has a broader meaning. For example, $\mathtt{animal}$ is a hypernym of $\mathtt{dog}$ since a dog is a type of animal.

\paragraph{Hyponym} A hyponym of a concept is another concept that has a more specific meaning. For example, $\mathtt{dog}$ is a hyponym of $\mathtt{animal}$.

\paragraph{Co-hyponym} Co-hyponyms of a concept are concepts that share the same hypernym with that concept. For example, $\mathtt{pembroke}$ is a co-hyponym of $\mathtt{golden\;retriever}$ as they both are hyponyms of the concept $\mathtt{dog}$.

% However, there are a few challenges to computing this value. Due to the absence of the oracle $\mathcal{O}$ in practice, we cannot determine $D_c$ and $D_{\bar{c}}$ given our image-only dataset $D$. Previous approach~\cite{bykov_labeling_2023} assumes that concept labels are available along with the dataset. Furthermore, even if we can approximate the oracle $\mathcal{O}$, it is non-trivial to find a dataset $D$ such that $D_c \neq \emptyset, \forall c$. Therefore, we introduce a procedure to generate a set of examples and non-examples $E_c$ and $E_{\bar{c}}$ as substitution for $D_c$ and $D_{\bar{c}}$ in Eq. (\ref{eq:score_validation}). 

\subsection{Sampling hard non-examples}
\label{sec:sub:sampling_hard}

Simply sampling random images to obtain the negative examples $E_{\Bar{c}}$ will not give us a good evaluation score $s(c)$ as random images might not contain enough concepts to distinguish the evaluated concept $c$ from other semantically similar concepts. For example, let's consider the scenario where the true concept $c=\mathtt{golden\;retriever}$, the candidate concept $c' = \mathtt{pembroke}$, and the set $E_{\Bar{c}}$ contains images of other \textit{non-dog} animals. Then, both $s(c)$ and $s(c')$ will likely be high since both concepts will result in a low activation for all samples in $E_{\Bar{c}}$. The reason is that the members of $E_{\Bar{c}}$ are both \textit{non}-$\mathtt{golden\;retriever}$ and \textit{non}-$\mathtt{pembroke}$ as they are all  \textit{non-dog}. As a result, a more careful sampling strategy to obtain  $E_{\Bar{c}}$ is needed to make a higher score more meaningful. 

In particular, we wish to generate $E_{\bar{c}}$ in a way that our score is high only when the neuron activates on that specific concept. To this end, our evaluation uses a concept's \textit{co-hyponyms} to generate $E_{\bar{c}}$. The intuition is illustrated in the previous scenario: the neuron not activating on $\mathtt{pembroke}$ is a stronger evidence for $c = \mathtt{golden\;retriever}$ than the neuron not activating on random images, given the similarity between $\mathtt{pembroke}$ and $\mathtt{golden\;retriever}$ (they are both breeds of dogs).

\subsection{Finding co-hyponyms with an LLM}
\label{sec:sub:finding_co_hyponyms}

For the first step of our validation process, given a concept $c$, we wish to find a set of its co-hyponyms to serve as strong non-examples for validation. While the task of finding co-hyponyms suggests the usage of a lexical database with hyponym relations built in like WordNet \cite{miller_wordnet_1994}, such operations are only possible for single concepts. For instance, humans can intuitively think of non-examples of ``red sedans" as ``blue sedans," yet such concepts are absent in WordNet. Instead, we leverage the flexibility of an LLM to generate co-hyponyms for all concepts. This reduces the need for ad-hoc methods and post-processing and enhances automation, which suits the goal of our method. 

To use an LLM for the task of finding co-hyponyms, we use a two-step process, inspired by Chain-of-Thought prompting~\cite{wei_chain--thought_2022}. First, given a concept $c$, we ask the LLM to find a hypernym of it. Then, in the same prompt, we ask it to generate concepts that share the same (previously generated) hypernym with $c$. In this setting, Chain-of-Thought prompting can improve performance by breaking down the complex problem into two smaller tasks. It also enables us to partially understand how the LLM come up with the answer.

There are two common failures the LLM can make. First, it can give a hypernym that is broader than what we want. For example, when asked for a hypernym of ``red sadans", the LLM tends to answer ``cars", instead of ``sedans of a particular color". Second, it might come up with a co-hyponym that is not entirely exclusive of the original concept. For instance, while it can correctly come up with the hypernym ``shapes" for the concept ``lines", it might use ``rectangles" as a co-hyponym. These behaviors can be attributed to the inherent limitations of LLMs having a limited understanding of concepts, compared to humans~\cite{shani_towards_2023}. Nevertheless, we alleviate these issues by using examples to trigger in-context learning~\cite{dong_survey_2023}. 

\subsection{From concepts to images}
\label{sec:sub:concept_to_image}

Now that we have the concept to be evaluated and its co-hyponyms, we wish to generate a set of examples and non-examples. We utilize a text-to-image diffusion model for this task. The diffusion model takes in a description of an image, which we will call a caption, and generates an image matching that caption. As this model expects a descriptive caption rather than a single word, we need to first convert the concepts to captions. For each concept, we want its captions to describe images containing that concept, but not its co-hyponyms. Likewise, we do not want the captions corresponding to the co-hypnoyms to unintentionally contain the original concept. To this end, we generate captions by considering pairs of concept and co-hyponym. For each pair, we explicitly ask the LLM to generate a pair of captions that contain one concept but not the other. Finally, the captions are passed to the diffusion model for image generation.

\begin{table}
    \centering
    \begin{tabular}{|c|c|c|}
        \hline
        Method & $D$ & Interpretable concepts \\
        \hline
        FALCON & ImageNet val & 11/100 \\
        Ours & ImageNet val & 99/100 \\
        \hline
    \end{tabular}
    \caption{Performance when labeling the first 100 neurons of the final layer of a ResNet50 trained on ImageNet.}
    \label{tab:imagenet_interpretable}
\end{table}

\begin{table}[b]
    \centering
    \begin{adjustbox}{width=\columnwidth,center}
    \begin{tabular}{|c|c|c|}
        \hline
        Unit & Ours & ImageNet class \\
        \hline
        0 & people holding fish & tench \\
        1 & goldfish & goldfish \\
        2 & sharks & great white shark \\
        3 & sharks & tiger shark\\
        4 & sharks & hammerhead\\
        5 & spotted creatures & electric ray\\
        6 & stingrays & stingray\\
        7 & red combs on chickens & cock\\
        8 & chickens & hen\\
        9 & flightless birds with long necks & ostrich\\
         \hline
    \end{tabular}
    \end{adjustbox}
    \caption{Descriptions of final layer neurons of a ResNet50 pre-trained on ImageNet.}
    \label{tab:imagenet_qualitative}
\end{table}

\section{Experiments}
\label{sec:experiments}

We conduct experiments to show the effectiveness of our method, comparing it to other baselines in terms of explanation quality. Section~\ref{sec:sub:implementation} provides details on our method's implementation and the experiment setting. Section~\ref{sec:sub:evaluate_accuracy} describes our experiment on ImageNet pre-trained models. In Section~\ref{sec:sub:evalute_clip}, we apply our method to the popular CLIP model and show the effectiveness of both concept discovery and concept validation. Due to the page constraint, we put more experiments, including ablation tests, more qualitative evaluations, a user study and experiments on different layers in the supplementary materials. The exact prompt used for the MLLM and the LLM will also be provided in the supp. mat.

\subsection{Implementation}
\label{sec:sub:implementation}

In this section, we provide the implementaion details of our method, including the specific model choice and the parameters used. In the subsequent experiments, the configurations are as follows unless otherwise noted. The source code will be made available.

For concept discovery, we use GPT-4V \cite{openai_gpt-4_2023} as the MLLM. To extract the exemplar representations, we let $\mu$ be equal to the $50$-th highest activation for each neuron, and sample $M=36$ images using (\ref{eq:clip_sampling}). The vision encoder $h$ is CLIP ViT-B/32 \cite{radford_learning_2021}.

For concept validation, we utilize LLaMA 2 13B \cite{touvron_llama_2023} as our LLM, both in co-hyponym generation and caption generation. To turn captions into images, we use Stable Diffusion XL Turbo \cite{sauer_adversarial_2023} with $\texttt{num\_inference\_steps}=4$ for enhanced prompt alignment. For each concept, we generate $5$ co-hyponyms, then, we generate $2$ pairs of captions for each pair of concept and co-hyponym. Finally, for each caption, we generate $5$ images, resulting in $|E_{\bar{c}}| = |E_c| = 5 \times 2 \times 5 = 50$. 

\begin{table}
    \centering
    \begin{adjustbox}{width=\columnwidth,center}
    \begin{tabular}{|c|c|c|c|}
        \hline
        Unit & FALCON & Ours & ImageNet class \\
        \hline
        48 & iguana,grass,face & reptiles & Komodo dragon \\
        63 & snake,head,carpet & snakes & Indian cobra \\
        70 & people,person,harvestman & spiders & harvestman \\
        71 & swatch,carpet,iii & scorpions & scorpion \\
        96 & parrot,yellow,head & toucans & toucan\\
        98 & water,baby,march & waterfowl & red merganser\\
         \hline
    \end{tabular}
    \end{adjustbox}
    \caption{Comparison of descriptions generated by FALCON and our method on final layer neurons of ResNet50 pre-trained on ImageNet, on cases where FALCON produced an explanation. }
    \label{tab:imagenet_vs_falcon}
\end{table}

\begin{figure}
    \centering
    \begin{subfigure}[b]{0.48\textwidth}
        \includegraphics[width=\textwidth]{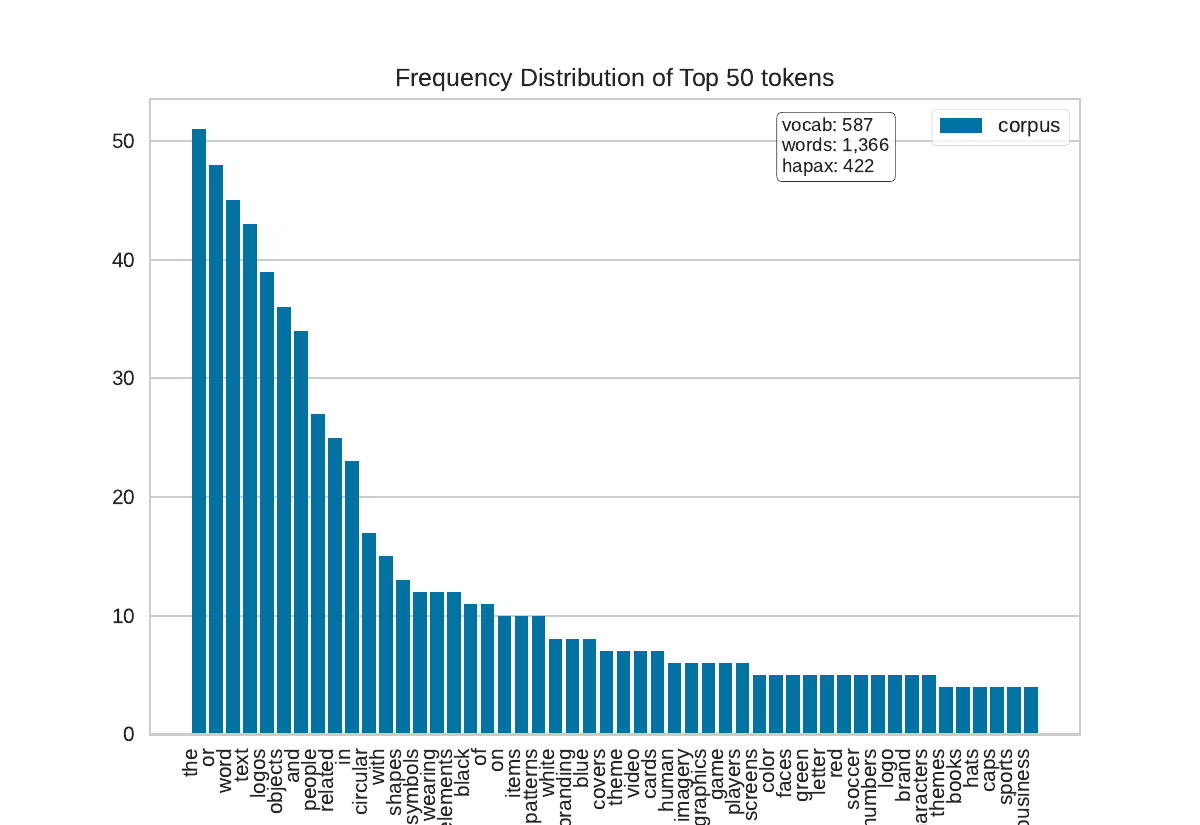}
        \caption{Ours}
    \end{subfigure}
    \centering
    \begin{subfigure}[b]{0.48\textwidth}
        \includegraphics[width=\textwidth]{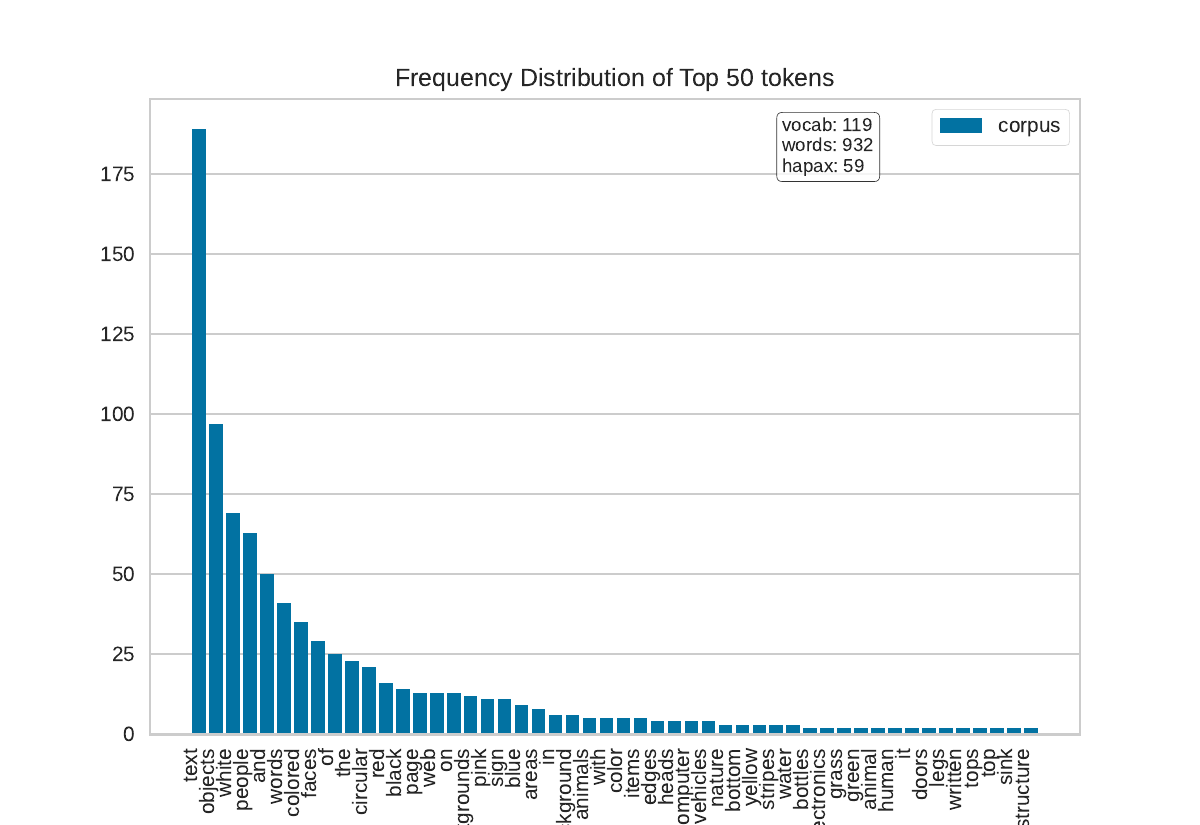}
        \caption{MILAN}
    \end{subfigure}
    \caption{Count of occurrences of the top-50 most frequent words in the explanations of the last layer of CLIP ResNet50.}
    \label{fig:clip_count}
\end{figure}

\subsection{Accuracy of explanations}
\label{sec:sub:evaluate_accuracy}

It is especially challenging to evaluate the accuracy of the generated explanations as we do not have the groundtruth concept in most cases. A common approach is evaluating on neurons whose expected behavior is given by the task, such as the final layer of classification models \cite{oikarinen_clip-dissect_2023,bykov_labeling_2023}. In this experiment, we compute explanations for the first 100 neurons of the final layer of ResNet50. We use the ImageNet-1K validation set as $D_{probe}$.  We compare our results with FALCON \cite{kalibhat_identifying_2023}, which we produce using their official code. Note that we do not compare with MILAN for this experiments because MILAN relies on thresholding the feature maps to obtain the exemplar representation, which the fully-connected layer we are explaining does not possess.

\begin{figure}
    \centering
    \includegraphics[width=0.48\textwidth]{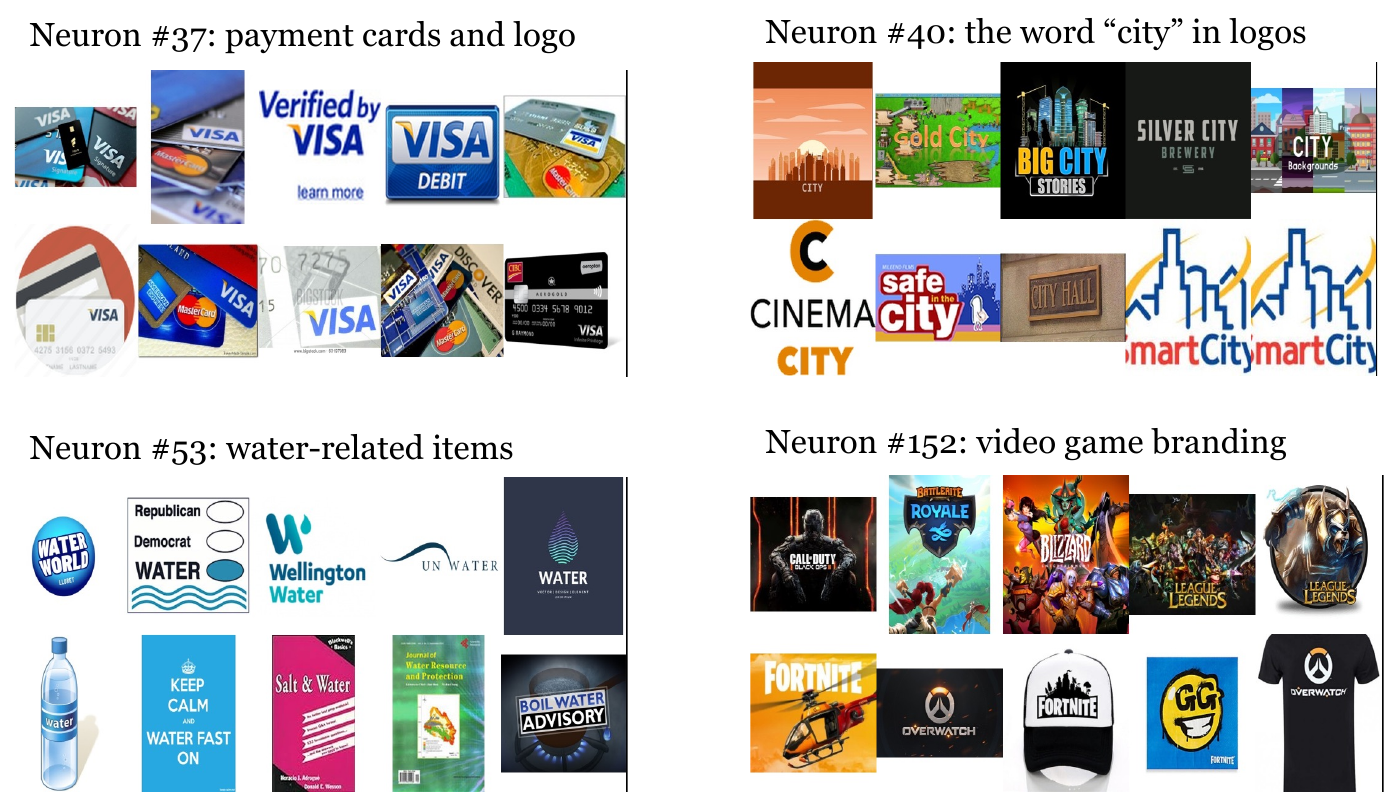}
    \caption{Examples of cases where our method produce a more detailed concept. All neurons are labeled as ``Text" by MILAN.}
    \label{fig:clip_text}
\end{figure}

Tables \ref{tab:imagenet_interpretable}, \ref{tab:imagenet_qualitative}, and \ref{tab:imagenet_vs_falcon} demonstrate our results. Note that we were able to interpret all but one out of the first 100 neurons, while FALCON was only able to recognize 11. From Table \ref{tab:imagenet_qualitative}, it can be seen that our method produces the exact ImageNet class, a superclass of it, or a related concept. Table 3 compares our descriptions with FALCON's top-3 words, in cases where they produced an explanation. While they were able to generate the exact concept in some case (unit 70), in general, our explanations are more consistently correct. 

% Note that since the output of a classifier is determined via comparisons of different activations (by passing through the softmax function), hence, within a neuron, it does not necessarily need to distinguish the target class from other classes (e.g., ``tiger shark" neuron can activate on other species of sharks).

% \begin{table}
% \begin{adjustbox}{width=\columnwidth,center}
% \begin{tabular}{r|lll}
% \multicolumn{1}{r}{}
% & \multicolumn{1}{l}{Heading 1}
% & \multicolumn{1}{l}{Heading 2}
% & \multicolumn{1}{l}{Heading 3} \\ \cline{2-4}
% Row 1 & Cell 1,1 & Cell 1,2 & Cell 1,3 \\
% Row 2 & Cell 2,1 & Cell 2,2 & Cell 2,3
% \end{tabular}
% \end{adjustbox}
% \end{table}

\subsection{Analyzing CLIP vision encoder}
\label{sec:sub:evalute_clip}

To show that our method can discover more insightful concepts on practical DNNs, we experiment with explaining the final layer of the backbone of a CLIP-ResNet50~\cite{radford_learning_2021}. To be precise, we generate explanations for the first 512 of the total 2048 neurons of the last layer (prior to attention pooling layer) of the backbone. We use a $1$M subset of LAION-400M~\cite{schuhmann_laion-400m_2021} as $D_{probe}$. In this experiment, we set $\mu$ to the $100$-th highest activation to account for a larger probing dataset. We compare our results with MILAN~\cite{hernandez_natural_2022}. For FALCON, their method generally only explains a small portion of neurons that they deem interpretable. Furthermore, this ratio depends on the choice of the threshold parameter. In contrast, our method and MILAN both aim to generate an explanation for an arbitrary neuron. As such, FALCON requires a different evaluation, which we provide in the supplementary material.

\subsubsection{Evaluating explanation variety }

\begin{figure}
    \centering
    \includegraphics[width=0.48\textwidth]{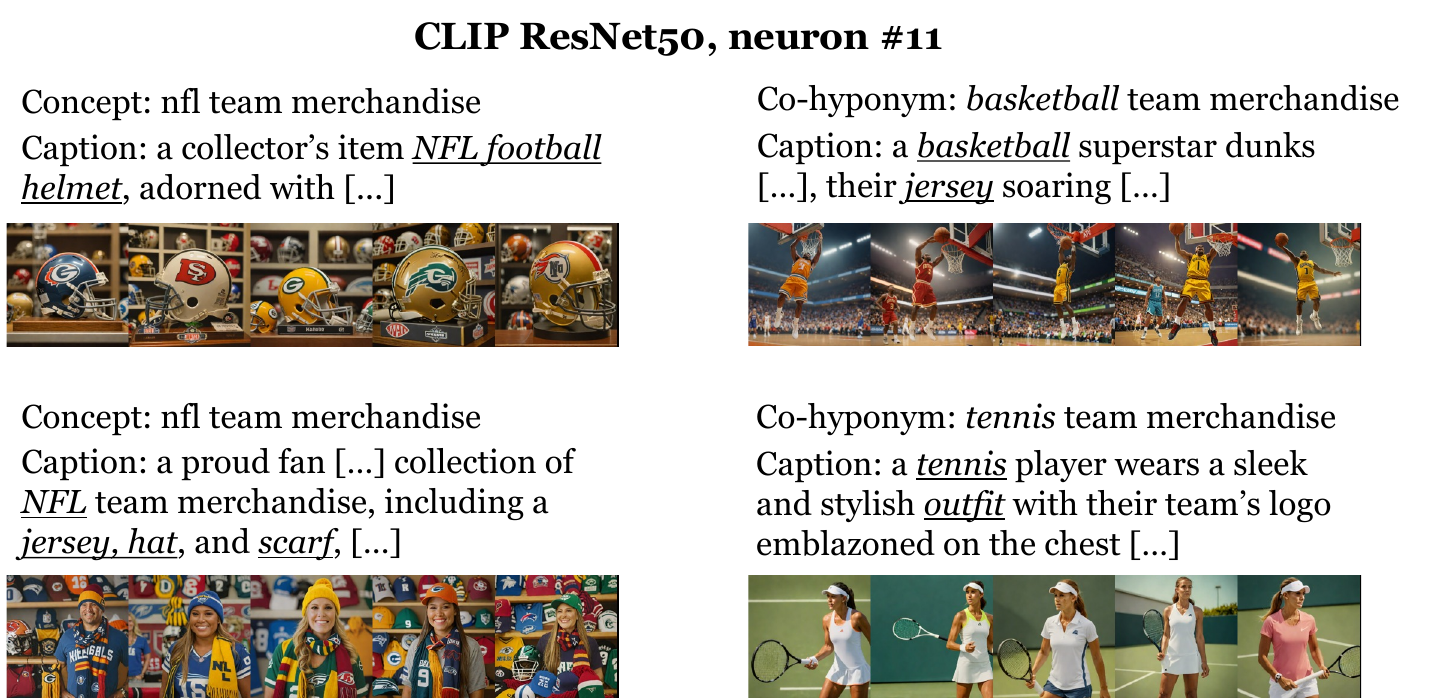}
    \caption{Example of concept validation in action. We show the process of evaluating our proposed concept ``nfl team merchandise", which results in a score of $0.984$ on the full set of $E_c$ and $E_{\bar{c}}$. We \textit{italicise} and \underline{underline} important words in the captions that are directly related to the concept or the co-hyponym. }
    \label{fig:evaluation_example}
\end{figure}

We analyze the generated explanations on a corpus level to show that our method generates more detailed concepts. In particular, we consider all explanations of each method as a corpus. Figure~\ref{fig:clip_count} illustrates the word-level analysis for each corpus. It counts the number of occurrences of each word in a corpus and plots the top 50 of them, as well as some other corpus-level statistics. Overall, we can see that the number of unique words (abbr. \textit{vocab}) and the number of words that appear only once (abbr. \textit{hapax}) are significantly higher than those of MILAN. This suggests that our method produce more specific concepts that is tailored to each neuron. Looking at the body of the distribution, MILAN labels neurons with generic concepts more frequently (``text": $175+$ usages versus ours' $50$, ``people": $50+$ vs. $30-$).

Observing that MILAN labels a significant $168$ out of $512$ neurons as ``text", we seek to compare those with our own explanations. In contrast with MILAN, we were able to extract a variety of different concepts. Fig. \ref{fig:clip_text} shows some examples of the results. In many cases, we discover the specific word on which the neuron activates. In other cases, we find the neuron activates on text coming from a particular subject, as in the ``video game" neuron. The results demonstrate that our method can automatically produce more specific concepts that provide more insights on the functionality of neurons.

\subsubsection{Explanation scores}

We demonstrate how the evaluation methodology proposed in Section~\ref{sec:concept_validation} works in practice. First, we show an example in Figure~\ref{fig:evaluation_example} with the concept to be evaluated, the co-hyponyms, the captions, and the set of examples and non-examples. Note that this is only a portion of the sets; we generate $100$ images in total for both examples and non-examples combined as described in Section~\ref{sec:sub:implementation}. We can observe that the exclusivity of the concept and its co-hyponyms was successfully transmitted to the captions and then to the images.

\begin{figure}
    \centering
    \includegraphics[width=0.48\textwidth]{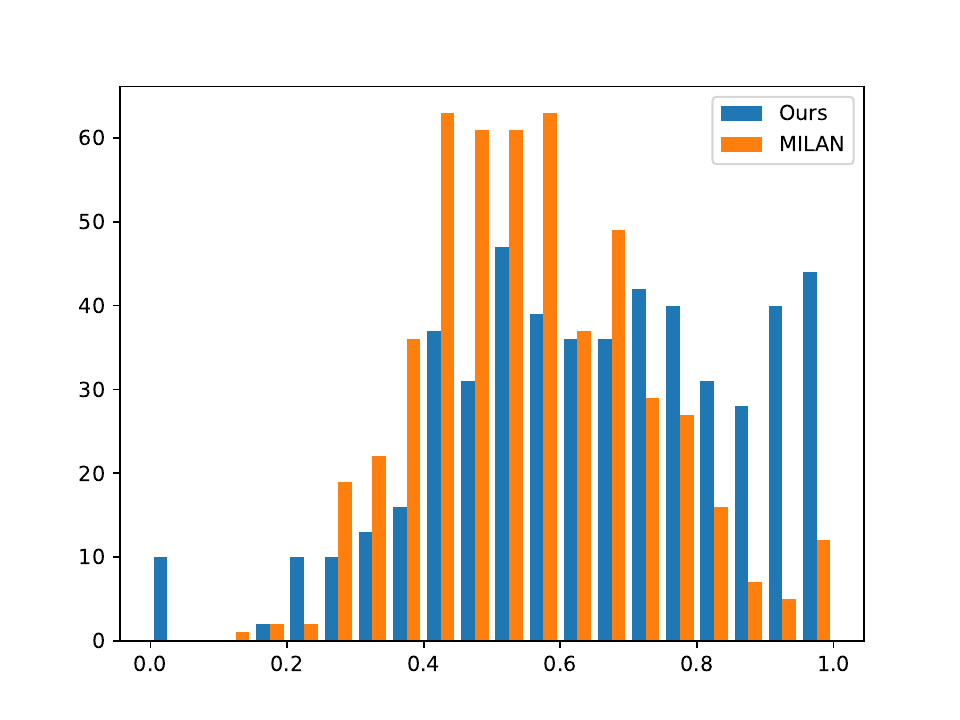}
    \caption{Distribution of explanation scores computed by our concept validation. The validated concepts are from explaining the last layer of CLIP ResNet50. Note that all cases where we get a zero score are from GPT-4V's refusal to answer. }
    \label{fig:clip_validation_score}
\end{figure}

We plot the distribution of the score for each neuron in Figure~\ref{fig:clip_validation_score}. The distribution suggests that our concepts better represents the chactersistic of the input that cause the neuron to fire strongly. Examples in Fig. \ref{fig:clip_text} supports this. We further evaluate our results with more qualitative examples and a user study, which we provide in the supplementary materials.

\section{Conclusion}
\label{sec:conclusion}

Previous methods for concept-based neuron explanations suffer from limited output language and require a curated dataset that defines the concepts. Our concept discovery method addresses this issue, leading to interpretable concepts that are more insightful and distinctive. To evaluate the generated concepts without any groundtruth, our evaluation method provides a meaningful to help users judge the quality of the result. Future work can seek to use this score as a signal to create a feedback loop, where the discovery method improves on its answer automatically. 

%% The file named.bst is a bibliography style file for BibTeX 0.99c
\bibliographystyle{named}
\bibliography{ijcai24}

\end{document}